\pdfoutput=1

\documentclass[11pt]{article}

\usepackage[preprint]{acl}

\usepackage{times}
\usepackage{latexsym}
\usepackage{hyperref}
\usepackage{array}

\usepackage[T1]{fontenc}

\usepackage[utf8]{inputenc}

\usepackage{microtype}

\usepackage{inconsolata}

\usepackage{graphicx}

\usepackage{CJKutf8}

\usepackage{booktabs}

\title{Multilingual Dialogue Generation and Localization with Dialogue Act Scripting}

 \author{
    \textbf{Justin Vasselli} \;\; 
    \textbf{Eunike Andriani Kardinata} \;\;\\
    \textbf{Yusuke Sakai} \;\;
    \textbf{Taro Watanabe} \\
    Nara Institute of Science and Technology
    \\
    \texttt{\{vasselli.justin\_ray.vk4, eunike.kardinata.ef9,}
    \\
    \texttt{ sakai.yusuke.sr9, taro\}@is.naist.jp}
    }
\begin{document}
\maketitle
\begin{abstract}
Non-English dialogue datasets are scarce, and models are often trained or evaluated on translations of English-language dialogues, an approach which can introduce artifacts that reduce their naturalness and cultural appropriateness.
This work proposes Dialogue Act Script (DAS), a structured framework for encoding, localizing, and generating multilingual dialogues from abstract intent representations. 
Rather than translating dialogue utterances directly, DAS enables the generation of new dialogues in the target language that are culturally and contextually appropriate.
By using structured dialogue act representations, DAS supports flexible localization across languages, mitigating translationese and enabling more fluent, naturalistic conversations.
Human evaluations across Italian, German, and Chinese show that DAS-generated dialogues consistently outperform those produced by both machine and human translators on measures of cultural relevance, coherence, and situational appropriateness.\footnote{Code and data available at \url{https://github.com/JVasselli/DialogueActScript}}
\end{abstract}

\section{Introduction}
Developing multilingual dialogue systems requires high-quality conversational data across diverse languages. However, authentic dialogue datasets are often scarce, costly, or difficult to obtain, making it challenging to train robust multilingual models~\citep{casanueva-etal-2022-nlu}. A common technique to build synthetic datasets is to generate dialogues by translating existing English datasets, but this approach often fails to capture cultural nuances and conversational norms, leading to two key issues: anglocentric biases, the assumption that English-speaking cultural contexts are universally applicable, and artifacts that make dialogues sound unnatural in the target language \citep{artetxe-etal-2020-translation}.

For instance, dialogues translated from English may retain American or British settings, mention culturally specific brands, or use names common in English-speaking countries but rare elsewhere. These issues may leave the dataset culturally English, limiting its usefulness for training and evaluating linguistically and culturally diverse dialogue systems.

To overcome these limitations, previous work has explored outline-based dialogue generation, where structured prompts rather than full English dialogues guide the creation of new conversational data \citep{shah-etal-2018-bootstrapping, majewska-etal-2023-cross}. \citet{majewska-etal-2023-cross} showed that this approach produces more natural and culturally appropriate dialogues than translations by professional human translators, as native speakers prefer localized adaptation over direct translation. However, their method relied on human annotators, limiting its scalability.

Building on this idea, we propose Dialogue Act Script (DAS), a structured framework for encoding, localizing, and generating multilingual dialogues. 
By abstracting conversations into intent-based representations before localization, DAS enables scalable, automatic adaptation of dialogue content while avoiding both anglocentric biases and translationese. This approach retains the strengths of outline-based annotation while leveraging large language models (LLMs) for both abstraction and localization, producing natural and culturally appropriate dialogues without requiring human annotation.

This work investigates the following research questions:

\begin{enumerate}
\item How accurately can DAS represent dialogue acts compared to human-created encodings?
\item To what extent does decoding from DAS preserve the meaning, fluency, and coherence of the original dialogue?
\item How well does DAS enable culturally appropriate localization?
\item Do dialogues generated with DAS yield higher-quality synthetic data than translations?
\item How does the three-step DAS pipeline compare with a single-step localization and translation prompt?

\end{enumerate}

By addressing these questions, we aim to demonstrate that DAS facilitates more culturally appropriate and coherent multilingual dialogue generation, as evaluated through both automated and human assessments across multiple languages.

To evaluate our approach, we use XDailyDialog~\citep{liu-etal-2023-xdailydialog} and the Cross-lingual Outline-based Dialogue (COD) dataset~\citep{majewska-etal-2023-cross}.
We compare DAS-generated dialogues against both machine-translated and human-translated versions of the original English dialogues. While translation is the most common method for building multilingual dialogue corpora, our results show that DAS-generated dialogues consistently outperform translation-based baselines on human evaluations.

\section{Related Work}
Translation-based methods are a common strategy for creating multilingual dialogue datasets \citep{mendonca-etal-2023-towards, anastasiou-etal-2022-machine, lin-etal-2021-xpersona, liu-etal-2023-xdailydialog}, but they can introduce issues that affect the quality of the downstream model. \citet{artetxe-etal-2020-translation} show that translated datasets fail to reflect naturally occurring multilingual data due to translation artifacts that distort linguistic patterns. These distortions can lead to unnatural exchanges and discourse inconsistencies, limiting their effectiveness for training conversational models.

To mitigate these issues, human-guided annotation methods have been explored. \citet{majewska-etal-2023-cross} introduced outline-based annotation, where human authors write dialogues into target languages from a localized outline of the original English dialogue. This approach enables cultural adaptation and prevents artificial alignment, leading to more natural multilingual dialogues. While effective, manual annotation is resource-intensive and difficult to scale.

An alternative is synthetic dialogue generation, where models generate dialogues autonomously. \citet{shah-etal-2018-bootstrapping} introduced Machines Talking to Machines (M2M) to generate large-scale synthetic dialogues, but such methods risk producing artificial conversational patterns that diverge from human interactions.

Recent work has explored how LLMs can generate structured representations from natural language. \citet{li-etal-2023-codeie} turned information extraction into a code generation task, using Code-LLMs to produce structured outputs. Similarly, \citet{sainz2024gollieannotationguidelinesimprove} introduced GoLLIE, a guideline-aware LLM for zero-shot IE, which uses annotation guidelines structured as Python classes to improve IE accuracy. These approaches show that LLMs can effectively generate structured, code-like representations as well as free-form text.

\section{Dialogue Act Script}

\subsection{Overview}
DAS is a structured framework for encoding dialogue through functional abstraction. It represents communicative intent using a predefined set of dialogue acts and parameters. Dialogue acts categorize utterances based on their communicative function (e.g., requesting, informing, or directing) rather than their surface form \citep{austin1962how}.

Rather than preserving the surface form of source-language dialogues through direct translation, DAS enables culturally adaptive generation by abstracting dialogues into structured intent representations and regenerating them in the target language.
This approach helps mitigate anglocentric biases, reduces artifacts associated with literal translation, and supports the creation of more natural, contextually appropriate dialogues across languages.

\begin{figure*}[t]
    \centering
    \includegraphics[width=\textwidth]{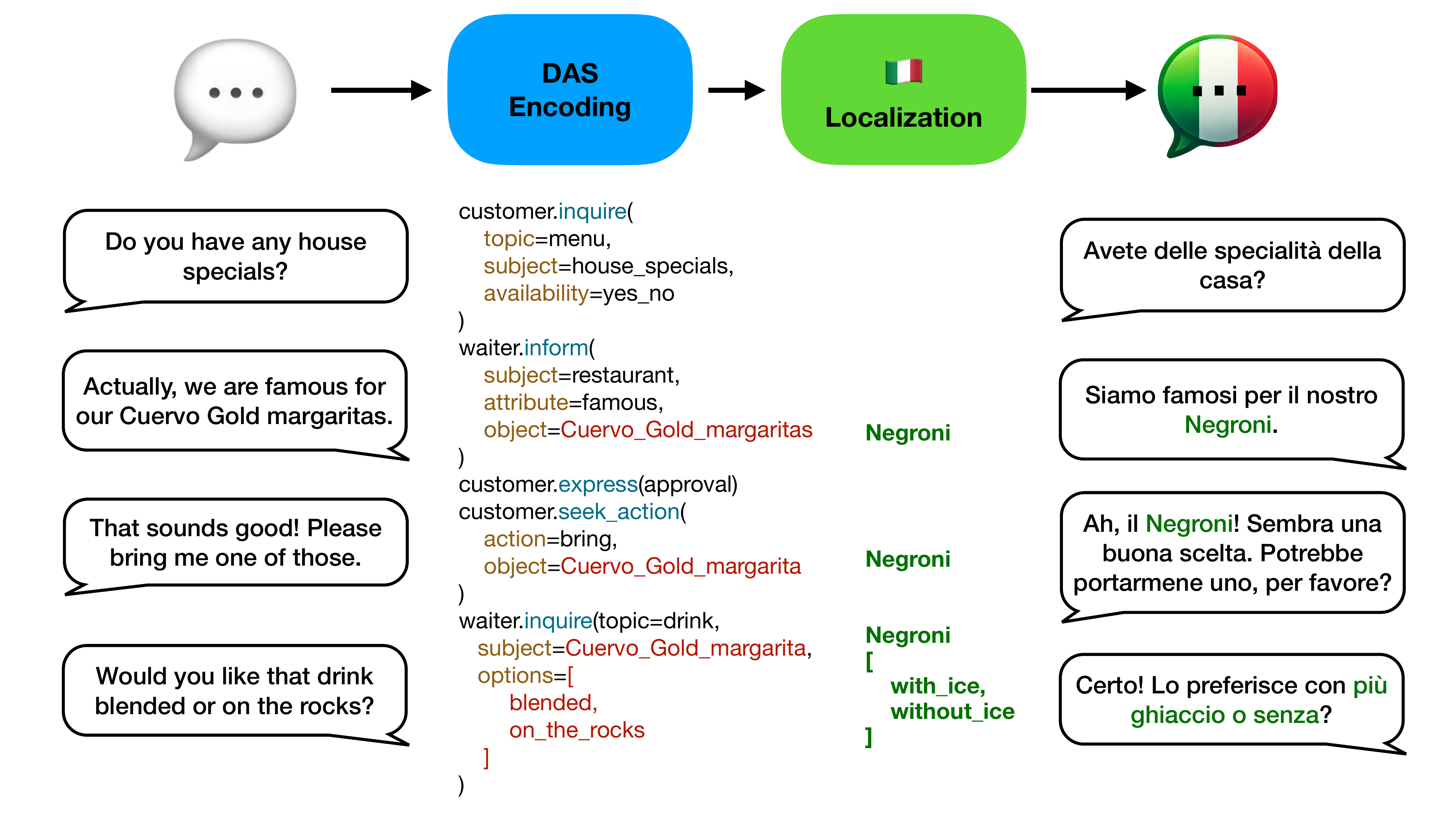}
    \caption{The DAS localization pipeline}
    \label{fig:flow}
\end{figure*}

\subsection{DAS Pipeline for Multilingual Dialogue Generation}
DAS facilitates the creation of multilingual dialogue data by culturally adapting dialogues through a three-step pipeline, as illustrated in Figure~\ref{fig:flow}:

\paragraph{Encoding:} Each utterance is converted into a DAS representation by classifying its dialogue act and extracting only the essential components needed to preserve its function, such as the speaker, action, relevant conditions, and timeframe. This structured abstraction preserves communicative intent while allowing for flexible multilingual reconstruction.
For example, the English utterance ``Actually, we are famous for our Cuervo Gold margaritas'' may be encoded as \texttt{inform(subject=restaurant, attribute=famous, object=Cuervo\_Gold\_margaritas)}.

\paragraph{Localization:} The DAS representation is then adapted to align with cultural norms in the target language by modifying relevant parameters (e.g., named entities, cultural references, or commonly used objects) while preserving the original dialogue act and intent. 
For instance, when adapting for an Italian audience, the drink \texttt{Cuervo\_Gold\_margaritas} might change to \texttt{Negroni}, reflecting a cocktail more commonly served in Italian bars.

\paragraph{Decoding:} Finally, the localized DAS representation is realized as fluent, coherent dialogue in the target language. This generation step reconstructs the conversation in a culturally appropriate manner while remaining faithful to the original communicative intent.
For example, the localized representation \texttt{inform(subject=restaurant, attribute=famous, object=Negroni)} could be decoded into Italian as: Siamo famosi per il nostro Negroni. (``We are famous for our Negroni'')

\subsection{Encoding}
The encoding process separates the form and content of an utterance, producing a structured representation that captures intent, dialogue acts, and semantic roles. This step consists of three key components:

\paragraph{Dialogue Act Classification:} Each utterance is assigned a dialogue act representing its communicative function (e.g., requesting, informing, expressing). This abstraction captures speaker intent independently of linguistic form, ensuring consistent representation across languages and phrasing styles. 
DAS is agnostic to the specific taxonomy used; any consistent set of communicative functions can be employed, or even omitted entirely in more free-form representations. 
In this study, we use a custom taxonomy developed to balance coverage, annotator consistency, and generation utility (see Section~\ref{sec:rq1}).

The schema was developed iteratively through human-in-the-loop refinement \citep{monarch2021human}: initial dialogue act categories were generated by prompting ChatGPT with example conversations, followed by a pilot human annotation phase. Categories with low inter-annotator agreement (e.g., \texttt{explain} were found to be difficult to distinguish from \texttt{inform} or \texttt{clarify}) were removed, and annotators were given the option to propose new dialogue acts when none of the existing ones fit. This process ensured that the final schema balanced flexibility with consistency while remaining informed by real conversational data.
For the full list of 15 dialogue acts in our annotation schema and corresponding examples, see Appendix~\ref{sec:functions}.

\paragraph{Slot Filling/Semantic Role Labeling:} Key roles and entities are assigned to fill the parameters of the dialogue acts. These parameters provide the minimum necessary information to reconstruct the utterance while preserving intent. This structured format ensures that critical details are explicitly captured, facilitating accurate localization and natural dialogue generation. For example, the utterance ``The wine list is on the second page of your menu.'' can be represented as: \texttt{inform(subject=wine\_list, location=second\_page, object=menu)} This representation captures the essential meaning while abstracting away language-specific phrasing, allowing for more flexible adaptation across different languages and cultural contexts.

\paragraph{Speaker Identification:} To maintain conversational coherence, each utterance is labeled with speaker roles. Speakers are typically identified as ``Speaker 1'' and ``Speaker 2,'' but when specific roles (e.g., ``Student'' and ``Teacher'') or named entities (``Susan'' or ``Billy'') are present, they are retained to enhance dialogue flow.

To capture broader conversational context, we prompted the model to generate scenarios with character biographies, allowing for greater consistency in tone and formality. These biographies included details such as names, ages, genders, and relationships between speakers to ground the dialogue in a more natural setting. Further details, including ablation studies, are provided in Appendix~\ref{appendix-bios}.

\subsection{Localization}
The localization step in DAS promotes cultural adaptability by enabling the generation of dialogues that are appropriate for the norms, entities, and expectations of the target language and culture.

In our implementation, localization is performed automatically by prompting a large language model to adapt the contextual frame (e.g., names, locations, and cultural references) and return an updated DAS dialogue by modifying relevant parameters (e.g., \texttt{location=New York} → \texttt{location=Beijing}) while preserving the underlying dialogue act. This allows the communicative function to remain consistent while the realization reflects culturally relevant details.

\subsection{Decoding}
Decoding involves generating natural-language dialogue from the DAS representation. Given the character descriptions and setting, which may have been localized, each DAS turn is realized as a fluent, contextually appropriate utterance in the target language.
This step also allows for fine-grained control over linguistic features. For example, developers can adjust the complexity or formality of the output to suit different audiences or use cases. A single DAS encoding such as \texttt{inquire(topic=menu, subject=house\_specials)} might be decoded with simple grammar and vocabulary (``Do you have house specials?''), or as a more formal version (``Would you be able to tell me about the house specials currently on offer?'') This flexibility makes DAS particularly useful for applications such as language learning where the output can be tuned to target different levels.

Decoding can be performed turn-by-turn (e.g., in interactive chatbot settings) or over the entire dialogue (e.g., for full-script localization). The approach is language-agnostic: once localized, a DAS representation can be realized in any language supported by the generation model. In our experiments, we evaluate decoding across Chinese, Italian, German, and English to assess DAS’s support for both cross-lingual and controlled-generation scenarios.

\section{Experiments}


To evaluate the effectiveness and flexibility of DAS, we conduct five experiments aligned with our research questions:

\textbf{RQ1:} Can LLMs reliably encode conversations into DAS representations?  
\textbf{RQ2:} Does the DAS representation preserve core meaning while allowing form variation?  
\textbf{RQ3:} How do slot-based localizations compare to human-annotated localizations?
\textbf{RQ4:} Can DAS localization produce dialogues that are more culturally relevant than direct translation?
\textbf{RQ5:} Does the modular DAS pipeline offer advantages over end-to-end prompting?  

For these experiments,  we selected 80 dialogues from the DailyDialog dataset~\citep{li-etal-2017-dailydialog}, which covers a range of conversational topics, lengths, and emotional tones.
To ensure a representative sample for translation and human evaluation, we applied the following criteria:
\begin{enumerate}
    \item \textbf{Conversation Length}: Dialogues with 8 to 16 turns were selected, resulting in an average of 10.92 turns per dialogue.
    \item \textbf{Topic Variety}: DailyDialog categorizes conversations into 10 distinct topics: Ordinary Life, School Life, Culture \& Education, Attitude \& Emotion, Relationship, Tourism, Health, Work, Politics, and Finance. We randomly selected 8 dialogues per topic to ensure diverse conversational contexts.
\end{enumerate}

We use the XDailyDialog dataset~\citep{liu-etal-2023-xdailydialog} as a reference for professionally translated dialogues in Italian, German, and Chinese. We also include a simple machine translation baseline by prompting GPT-4o to translate directly from English (see Appendix~\ref{mt-prompt} for the prompt).

While DAS is flexible and can be applied with different models at each stage, in this study, we use GPT-4o (gpt-4o-2024-08-06) and GPT-4o-mini (gpt-4o-mini-2024-07-18) for encoding, localization, and decoding \footnote{GPT models were accessed through OpenAI's API and followed OpenAI's terms for API usage. The number of parameters of these models is undisclosed. We spend approximately \$100 USD on experiments.}.  Temperature was set to 0 for encoding to ensure consistent DAS representations across runs, as variation in function labeling could affect reproducibility. For localization and decoding, a temperature of 0.2 was chosen to allow for natural variation in expression while still preserving core meaning.

\subsection{RQ1 - Encoding Accuracy}
\label{sec:rq1}
To assess the reliability of DAS function annotations, we conducted an inter-annotator agreement (IAA) study comparing human-human consistency and human-GPT agreement for DAS function labeling.
Two native English speakers labeled 105 dialogue turns from five randomly selected conversations, using a predefined set of DAS functions. Rather than adopting an existing taxonomy, we designed a new, task-specific schema to test how well large language models could apply unfamiliar classification schemes.
This choice also reduced the risk of data leakage, since widely used taxonomies may have been encountered during model training. Annotators received the same function definitions and examples as the language models, ensuring consistent guidelines. The full taxonomy and examples are provided in Appendix~\ref{sec:functions}. We evaluated GPT-4o and GPT-4o-mini using identical prompts and instructions. The results are shown in Table~\ref{tab:IAA-results}.

\begin{table}[t]
    \centering
    \small
    \begin{tabular}{l|cccc}
        \toprule
        \textbf{Annotator} & \textbf{Human1} & \textbf{Human2} & \textbf{GPT4o-mini}\\
        \midrule
        \textbf{Human2} & 0.844 & - & - \\ 
        \textbf{GPT4o-mini} & 0.765 & 0.746 & -\\
        \textbf{GPT4o} & 0.822 & 0.769 & 0.805 \\
        \bottomrule
    \end{tabular}
    \caption{Inter-annotator agreement (Cohen's kappa) results for DAS function annotation.}
    \label{tab:IAA-results}
\end{table}

High agreement between human annotators ($\kappa$ = 0.844) suggests that the schema supports annotation consistency. Substantial agreement between humans and GPT-4o ($\kappa$ = 0.822, 0.769) indicates that the LLM can reliably apply dialogue acts when provided with clear definitions and examples. GPT-4o-mini also maintained reasonable agreement ($\kappa = 0.765$, $0.746$), though slightly lower than GPT-4o.


To assess the compatibility of DAS encoding with existing dialogue act schemes, we conducted an additional experiment using the DailyDialog taxonomy (Inform, Question, Directive, Commissive). GPT-4o was prompted to assign one of these four acts to each turn. 
GPT-4o achieved high F1 scores for Inform (0.92) and Question (0.94), which together covered 87.9\% of all turns. Performance on the comparatively rarer Directive (0.63) and Commissive (0.64) was lower. This suggests that GPT-4o is strong at classifying more common and straightforward dialogue acts.

\subsection{RQ2 - Encoding Meaning Preservation}
\label{sec:back-to-english}
To assess how well DAS preserves meaning while allowing for structural changes, we decoded DAS-encoded English dialogues back into English and compared them to the original dialogues. This evaluation serves two key purposes: first, to determine whether DAS retains the essential communicative intent of a conversation, and second, to examine whether DAS reconstruction introduces meaningful paraphrasing effects that could be useful for fluency enhancement or synthetic data generation.

We conducted human assessments using a pair of native English speakers. Annotators were shown pairs of conversations, the original dialogue and its DAS-decoded version, and asked the following questions:
\begin{enumerate}
    \item Fluency: Which conversation has the more fluent or natural sounding language?
    \item Coherence: Which conversation makes the most logical sense? (No sudden changes of topic, each turn naturally follows the previous on)
    \item Situational Appropriateness: Which conversation has the more appropriate tone or style for the situation?
    \item Meaning Preservation: How similar are the conversations in meaning?
\end{enumerate}

For the first three questions, annotators were allowed to choose A, B, Both, or Neither. Win rates were calculated by assigning a point to a system each time it was chosen over another or when ``Both'' was selected; no points were awarded when ``Neither'' was selected. Meaning preservation was reported on a Likert scale, with 1 indicating the conversations had completely different meanings, and 5 being they were identical in meaning.

\begin{table}[t] 
\centering 
\small 
\begin{tabular}{lccc} 
\toprule
\textbf{Metric} & \textbf{DAS} & 
\textbf{Original} \\ 
\midrule
Fluency & 0.727 & 0.455 \\
Logical Flow & 1.000 & 0.636 \\
Situational & 0.909 & 0.636 \\
\midrule 
Meaning Preservation & \multicolumn{3}{c}{Avg. Score: 4.63/5} \\
\bottomrule
\end{tabular} 
\caption{Human evaluation of DAS-decoded English compared to the original dialogues.} 
\label{tab:human-eval-back2eng} 
\end{table}

The results, reported in Table~\ref{tab:human-eval-back2eng}, suggest that DAS decoding does not introduce many disfluencies or disrupt conversational flow. In most cases, DAS produces output that is at least as coherent and appropriate as the original dialogue, with notable improvements in fluency for over half of the conversations.
The high meaning preservation score (4.63/5) indicates that DAS retains core intent effectively, even when rewording utterances. Although DAS generally improved fluency, situational appropriateness was slightly lower in some cases, suggesting that certain stylistic nuances may change during decoding.

In addition to human evaluation, we used automated metrics to assess the semantic similarity and structural differences between the original dialogues and their DAS-decoded versions.  See Appendix \ref{back-to-english-auto} for details and results of this experiment.

\subsection{RQ3 - Cultural Adaptation}

To evaluate whether the DAS localization process produces culturally adapted slot substitutions similar to those made by human annotators, we conducted a slot-level comparison using dialogues from the Cross-lingual Outline-based Dialogue (COD) dataset~\citep{majewska-etal-2023-cross}.

COD was created through manual rewriting of outlines, including a localization step where culturally specific named entities were replaced by native speakers. We applied DAS localization to the 92 original English dialogues from the COD development data and evaluated the 1196 annotated slots that contain values. First, we look at how well DAS identifies slots that should be changed. We calculate the F1 score for localized slots using COD as the gold standard and report the results in Table~\ref{tab:slot-eval}.

\begin{table}[t]
\centering
\small
\begin{tabular}{lccc}
\toprule
\textbf{Language} & \textbf{Precision} & \textbf{Recall} & \textbf{F1} \\
\midrule
Arabic & 0.929 & 0.760 & 0.836 \\
Indonesian & 0.865 & 0.802 & 0.832 \\
Russian & 0.894 & 0.796 & 0.843 \\
Swahili & 0.852 & 0.703 & 0.770 \\

\midrule
Average & 0.885 & 0.765 & 0.820\\
\bottomrule
\end{tabular}
\caption{Slot-level comparison between GPT-localized and human-localized dialogues.}
\label{tab:slot-eval}
\end{table}

The results show that DAS achieves consistently high precision across languages, suggesting it rarely substitutes slots unnecessarily. Recall is lower, particularly for Swahili, meaning the system sometimes misses substitutions that human annotators considered important. Overall, the average F1 of 0.82 indicates that DAS captures most of the cultural adaptations present in COD while erring on the side of conservatism, preserving original values unless confident in an appropriate substitute.



\begin{figure*}[t]
    \centering
    \includegraphics[width=\textwidth]{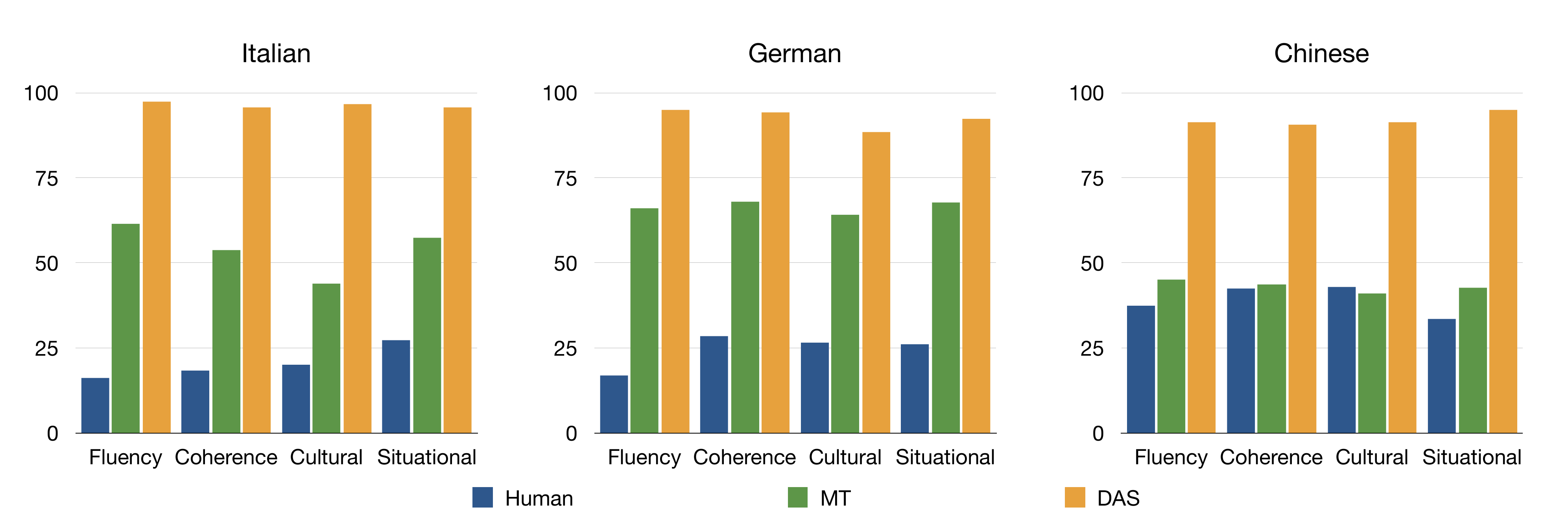}
    \caption{Win rates of each system across evaluation criteria (fluency, coherence, cultural relevance, and situational appropriateness). Higher win rates indicate stronger performance in pairwise comparisons.}
    \label{fig:win-rates-closed}
\end{figure*}




\begin{table}[t]
\centering
\small
\begin{tabular}{lcc}
\toprule
\textbf{Slot type} & \textbf{Correct / Total} & \textbf{Accuracy} \\
\midrule

Airline      & 25 / 26 & 0.96 \\
Airport      & 2 / 4   & 0.50 \\
Attraction   & 15 / 16 & 0.94 \\
City Name    & 66 / 66 & 1.00 \\
Given Names   & 20 / 20 & 1.00 \\
Music Album   & 15 / 22 & 0.68 \\
Music Artist  & 17 / 26 & 0.65 \\
Movie Actor   & 6 / 9   & 0.67 \\
Movie Director & 7 / 10  & 0.70 \\
Movie Title  & 16 / 24 & 0.67 \\
Song Title   & 29 / 38 & 0.76 \\
\bottomrule
\end{tabular}
\caption{Manual annotation of Indonesian DAS outputs by slot type.}
\label{tab:indo-slots}
\end{table}

Manual annotation of Indonesian outputs (Table~\ref{tab:indo-slots}) shows that localization accuracy is near-perfect for broad cultural categories such as city names, airlines, attractions, and given names, but weaker for domain-specific entities like music artists or movie titles. This suggests that DAS reliably handles common, geographically grounded references but is less consistent when deeper domain knowledge is required.

\subsection{RQ4 - Decoded Dialogue Quality}
\label{human-evaluation}

\begin{table*}[t]
    \centering
    \begin{tabular}{l|ccccccc}
    \toprule
      Aspect & DAS Win \% & Both \% & Neither \% & Human Win \% & DAS WR & Human WR & $p$\\
      \midrule
      \textbf{Italian} & & & & \\
      Fluency & 84.6 & 12.3 & \phantom{0}0.0 & \phantom{0}3.1 & 96.9 & 15.4 & <0.001 \\
      Coherence & 86.9 & \phantom{0}7.7 & \phantom{0}0.0 & \phantom{0}5.4 & 94.6 & 13.1 & <0.001\\
      Cultural & 85.4 & 10.8 & \phantom{0}0.8 & \phantom{0}3.1 & 96.2 & 13.8 & <0.001\\
      Situational & 85.4 & \phantom{0}9.2 & \phantom{0}0.0 & \phantom{0}5.4 & 94.6 & 14.6 & <0.001\\
        \midrule
      \textbf{German} & & & \\
      Fluency & 81.5 & 13.8 & \phantom{0}3.1 & \phantom{0}1.5 & 95.4 & 15.4 & <0.001 \\
      Coherence & 83.8 & 13.1 & \phantom{0}1.5 & \phantom{0}1.5 & 96.9 & 14.6 & <0.001\\
      Cultural & 71.5 & 17.7 & \phantom{0}8.5 & \phantom{0}2.3 & 89.2 & 20.0 & <0.001\\
      Situational & 72.3 & 20.8 & \phantom{0}1.5 & \phantom{0}5.4 & 93.1 & 26.2 & <0.001\\
      \midrule
      \textbf{Chinese} & & & \\
      Fluency & 75.3 & 13.5 & \phantom{0}1.1 & 10.1 & 88.8 & 23.6 & <0.001 \\
      Coherence & 75.3 & 14.6 & \phantom{0}1.1 & \phantom{0}9.0 & 89.9 & 23.6 & <0.001\\
      Cultural & 68.5 & 19.1 & \phantom{0}3.4 & \phantom{0}9.0 & 87.6 & 28.1 & <0.001\\
      Situational & 76.4 & 16.9 & \phantom{0}2.2 & \phantom{0}4.5 & 93.3 & 21.3 & <0.001\\
    \end{tabular}
    \caption{Pairwise evaluation of DAS generated conversations vs human translated conversations. Ties were classified into ``Both'' and ``Neither''. Win Rate (WR) is Win + ``Both''. $p$-value calculated through two-tailed binomial test}
    \label{tab:das-vs-human-detailed}
\end{table*}

\begin{table*}[t]
    \centering
    \begin{tabular}{l|ccccccc}
    \toprule
      Aspect & DAS Win \% & Both \% & Neither \% & SP Win \% & DAS WR & SP WR & $p$\\
      \midrule
      \textbf{Italian} & & & & \\
      Fluency & 62.5 & 32.5 & \phantom{0}0.0 & \phantom{0}5.0 & 95.0 & 37.5 & <0.001\\
      Coherence & 75.0 & 13.8 & \phantom{0}0.0 & 11.2 & 88.8 & 25.0 & <0.001\\
      Cultural & 62.5 & 27.5 & \phantom{0}0.0 & 10.0 & 90.0 & 37.5 & <0.001\\
      Situational & 75.0 & 17.5 & \phantom{0}0.0 & \phantom{0}7.5 & 92.5 & 25.0 & <0.001\\
        \midrule
      \textbf{German} & & & \\
      Fluency & 62.5 & 21.2 & \phantom{0}0.0 & 16.2 & 83.8 & 37.5 & <0.001\\
      Coherence & 57.5 & 26.2 & \phantom{0}0.0 & 16.2 & 83.8 & 42.5 & <0.001\\
      Cultural & 60.0 & 22.5 & \phantom{0}0.0 & 17.5 & 82.5 & 40.0 & <0.001\\
      Situational & 67.5 & 12.5 & \phantom{0}0.0 & 20.0 & 80.0 & 32.5 & <0.001\\
      \midrule
      \textbf{Chinese} & & & \\
      Fluency & 53.8 & 25.0 & \phantom{0}0.0 & 21.2 & 78.8 & 46.2 & \phantom{<}0.001 \\
      Coherence & 61.3 & 13.8 & \phantom{0}0.0 & 25.0 & 75.0 & 38.8 & <0.001\\
      Cultural & 55.0 & 27.5 & \phantom{0}0.0 & 17.5 & 82.5 & 45.0 & <0.001\\
      Situational & 67.5 & \phantom{0}8.8 & \phantom{0}0.0 & 23.8 & 76.2 & 32.5 & <0.001\\
    \end{tabular}
    \caption{Pairwise evaluation of DAS generated conversations vs Single Prompt (SP) translated+localized conversations. Ties were classified into ``Both'' and ``Neither''. Win Rate (WR) is Win + ``Both''. $p$-value calculated through two-tailed binomial test}
    \label{tab:pipeline_vs_single}
\end{table*}
To assess the quality of dialogues generated through DAS localization, we conducted a human evaluation on the generated target-language text. We compared the DAS generated text to two different translation baselines, as translation is the common technique for generating multilingual datasets.

Two native speakers each of Chinese, Italian, and German\footnote{See Appendix~\ref{human-annotation} for more detailed information on the human annotators} were recruited to compare DAS-localized dialogues against two baselines: human-translated dialogues from the XDailyDialog dataset, and machine-translated dialogues generated by prompting GPT-4o to directly translate the English source. Although both baselines involve translation, we do not evaluate ``translation accuracy''; instead, we treat these as standard approaches to multilingual dialogue generation and compare them to DAS as alternative generation methods.
As we are not judging typical translation metrics such as fidelity to the source, we do not show the annotators the original English dialogues.

Annotators were presented with a random pair of generated dialogues and asked the following questions:
\begin{enumerate}
    \item Fluency: Which conversation has the more fluent or natural sounding language?
    \item Coherence: Which conversation makes the most logical sense? (No sudden changes of topic, each turn naturally follows the previous on)
    \item Cultural Relevance: Which conversation feels more culturally (Italian/German/Chinese)?
    \item Situational Appropriateness: Which conversation has the more appropriate tone or style for the situation?
\end{enumerate}

Each annotator was allowed to select A, B, Both, or Neither for each question. Win rates are calculated as in section \ref{sec:back-to-english}, with ``Both'' counted as a win for both systems, and ``Neither'' counting as a loss for both. 

The results, shown in Figure~\ref{fig:win-rates-closed}, demonstrate that DAS consistently outperforms or matches both machine translation and human translations, particularly in cultural relevance and situational appropriateness. To assess statistical significance, we conducted a two-tailed binomial test, comparing wins and losses only (excluding ``Both'' and ``Neither'' responses). Across all three languages and all four evaluation criteria, DAS was preferred over the professional translations with high significance ($p < 0.001$). Table \ref{tab:das-vs-human-detailed} shows the detailed breakdown of annotations between pairwise judgments of DAS generated conversations and human translated conversations.

While the lower performance of the professionally translated dialogues may seem surprising at first, these results may simply reflect a fundamental difference in goals between traditional translation workflows and open-ended, culturally adaptive dialogue generation.
Professional translators often aim to preserve the original meaning as faithfully as possible. However, as we saw in ~\ref{sec:back-to-english}, the original dialogues contain  disfluencies, inconsistent tenses, or informal phrasing, all of which could have led to translations that feel rigid or unnatural in the target language. For example, one annotator noted that a professional translation shifted awkwardly between past and present tense, likely due to literal adherence to the original English. Such artifacts, while arguably accurate, are often dispreferred by native speakers evaluating fluency and conversational naturalness.

In contrast, GPT-4o, even under a simple translation prompt, tends to ``clean up'' awkward or inconsistent source material during generation, resulting in smoother target-language output. DAS goes a step further by discarding the surface form of the source entirely. Its reliance on abstract, intent-based representations allows for even greater flexibility in how conversations are realized, enabling shifts in style, tone, and cultural framing that better align with local conversational norms. For the details of the pairwise annotation between DAS generated conversations and GPT-4o translated conversations, please see Table \ref{tab:das-vs-mt-detailed} in the Appendix.

It is also important to consider the nature of the evaluation setup as a pairwise comparison instead of quality scores. As such, the fact that professional translations were often dispreferred does not imply that they are low-quality. Instead, it reflects their performance relative to more adaptive systems in a specific conversational context.

These findings align with those reported by \citet{majewska-etal-2023-cross}, who similarly observed that dialogue outputs generated from abstract representations were preferred over direct translations. Together, these results suggest that abstraction-based pipelines like DAS may be more effective than form-preserving translation approaches when the goal is to generate fluent, culturally appropriate dialogue, rather than to maintain strict fidelity to source-language wording.

\subsection{RQ5 - DAS Pipeline vs. Single Prompt}
\label{sec:rq5}

As the DAS pipeline currently relies on GPT-4o for all three steps, a natural question arises: could a single prompt accomplish the same task more efficiently? To test this, we constructed a baseline that prompts GPT-4o to directly translate and localize the English dialogue into the target language in one step. This prompt uses the localization instruction used in the DAS pipeline but skips the intermediate abstraction step entirely.






As shown in Table~\ref{tab:pipeline_vs_single}, despite receiving the same high-level localization instructions, the single-prompt baseline consistently underperformed across all evaluation criteria. Human annotators noted several recurring issues with the single-prompt approach. In many cases, cultural localization was incomplete or entirely absent. For example, in the example illustrated in Figure~\ref{fig:flow}, references to ``Cuervo Gold margaritas'' were preserved verbatim rather than adapted to locally appropriate alternatives. Annotators also reported that the single-prompt outputs tended to sound ``textbook-like'' or sometimes inappropriately casual or formal. In particular, one Italian annotator described the style as stiff and lacking conversational naturalness. 


These results demonstrate that the performance gains observed with DAS are not solely due to the use of GPT-4o, but emerge from the modular pipeline itself. Explicitly separating the localization and decoding steps appears to improve both cultural relevance and fluency, even when using the same base model.

\section{Conclusion}

This study introduced Dialogue Act Script, a modular framework for abstracting and localizing multilingual dialogues through intent-based representations. By separating the processes of encoding, localization, and decoding, DAS enables explicit cultural adaptation and flexible realization of dialogue across languages.

In our experiments, DAS generated synthetic dialogues consistently outperformed both human and machine translations. As shown in Section~\ref{sec:rq5}, these gains reflect the benefits of modular design: separating communicative intent from surface form enables more flexible and culturally adaptive generation, independent of any single model like GPT-4o.

A central strength of DAS is its modularity. Each step in the pipeline is independent, allowing for greater adaptability. While this paper used GPT-4o for all stages, there is growing evidence of cultural and stylistic biases in LLMs, including anglocentric tendencies and uneven performance across languages \citep{naous-etal-2024-beer}. DAS makes it possible to substitute any component with an alternative model, a retrieval-based method, or a human-in-the-loop process. Exploring these modular configurations is a promising direction for future work.

Beyond localization, DAS presents new opportunities for synthetic data generation, multilingual AI training, and rule-based machine translation in low-resource settings. We leave addressing challenges such as annotation consistency, scalability, and domain adaptability to future work.

\section*{Limitations}
Several limitations apply to the current version of this work.
First, the DAS pipeline relies on multiple calls to LLMs, which increases computational cost. Although the DAS encoding step is reusable across languages, deploying the pipeline in low-resource or compute-constrained environments remains challenging. Future work should explore lighter-weight or retrieval-based alternatives for each step of the pipeline, especially for localization and decoding.

Second, our human evaluation is limited to the XDailyDialog dataset, which consists of open-domain chitchat dialogues. While this setting is useful for evaluating conversational fluency and cultural adaptation, it does not represent the structure or communicative goals of more specialized domains. Future work should explore how well DAS generalizes to task-oriented or domain-specific dialogues, such as those found in customer support, healthcare, or legal contexts.

Third, while DAS is designed to enable cultural adaptation, the current implementation relies entirely on GPT-4o for all steps of the pipeline. This raises valid concerns about inherited cultural biases from the underlying model, particularly given prior findings on anglocentric bias in LLMs \citep{naous-etal-2024-beer}. Our intention is not to claim that GPT-4o is an ideal solution for localization, but rather to evaluate whether DAS, as an abstraction framework, enables more flexible and culturally responsive generation than translation alone. DAS is modular by design: each step can be implemented independently. The localization step, in particular, does not require generation and could be replaced with rule-based substitutions, retrieval systems, or human annotations. We see improving the localization step as an important direction for future work.

Fourth, while we evaluate the end-to-end quality of localized dialogues through pairwise human judgments, we do not directly validate the cultural appropriateness of individual slot substitutions. A more targeted evaluation of the localization step, for instance through native speaker judgments of entity familiarity or cultural fit, remains an important area for future study.

Finally, our evaluation primarily targets well-resourced languages such as Chinese, Italian, and German. The performance of DAS in low-resource or morphologically complex languages remains uncertain. Although we include slot-level analysis for additional languages in the COD dataset, further work is needed to understand how DAS performs in settings where LLMs have limited coverage or cultural knowledge.

\section*{Ethical Considerations}

As with all work involving LLMs, our framework inherits risks related to unintended social and cultural biases. One recurrent pattern was a default tendency to assign male-female gender roles to dialogue participants, with 88\% of conversations exhibiting this distribution. Although some mitigation strategies were attempted, this bias persisted. We did not conduct an exhaustive analysis of other cultural or representational biases, particularly in localized content. Future work should include more targeted bias evaluation and mitigation strategies, and we caution users of DAS to critically assess outputs, especially in real-world or sensitive applications.

The use of LLMs in our pipeline contributes to the environmental footprint of large-scale NLP systems. Future work could explore lightweight models or optimization strategies to improve the sustainability of multilingual generation frameworks like DAS.

We use the XDailyDialog dataset under the Apache-2.0 License, and its base dataset, DailyDialog, under CC BY-NC-SA 4.0. Both licenses permit research use with attribution. The original English conversations were sourced from websites for English learners and primarily reflect informal chitchat dialogues, which may not generalize to other conversational domains.

While DAS supports cultural adaptation of dialogues, it is not designed for high-stakes applications such as legal, medical, or financial translation. Any deployment beyond research settings should include human validation and safeguards to ensure responsible use.

AI tools such as ChatGPT and GitHub Copilot were used for minor language revisions and line-level code assistance, but all research design and outputs were authored and verified by the research team.

\bibliography{anthology,custom}

\appendix

\section{DAS Functions}
\label{sec:functions}
\begin{enumerate}
    \item \textbf{Inquire}  
    \subitem Seeks information or clarification. Includes direct questions or indirect inquiries.  
    \subitem \textit{What time does the meeting start?}  

    \item \textbf{Clarify}  
    \subitem Seeks to resolve ambiguity, misunderstanding, or confusion in a previous statement. Often involves rephrasing, elaboration, or highlighting specific details.  
    \subitem \textit{I meant next Tuesday.}  

    \item \textbf{Inform}  
    \subitem Provides factual information, details, or observations.  
    \subitem \textit{This policy was updated last week.}  

    \item \textbf{Express}  
    \subitem Communicates emotions, attitudes, or subjective opinions.  
    \subitem \textit{That's an excellent idea!}  

    \item \textbf{Agree}  
    \subitem Affirms or aligns with a previous statement.  
    \subitem \textit{Yeah, that makes sense to me.}  

    \item \textbf{Disagree}  
    \subitem Explicitly communicates disagreement or contradiction with a previous statement or idea. May provide reasoning or counterarguments but does not necessarily imply hostility or conflict.  
    \subitem \textit{That doesn't seem right to me.}  

    \item \textbf{Commit}  
    \subitem Explicitly agrees or promises to take a future action, either in response to a request or as a declaration of intent. The action must be something the speaker is directly responsible for performing.  
    \subitem \textit{Yes, I’ll take care of that.}  

    \item \textbf{Acknowledge}  
    \subitem Neutral receipt of information, often used for backchanneling or minimal responses.  
    \subitem \textit{I see.}  
    \subitem \textit{Okay.}  

    \item \textbf{Seek Action}  
    \subitem Represents any utterance where the speaker seeks to influence the listener’s behavior, encompassing both polite requests and authoritative commands.  
    \subitem \textit{Could you please send me the file?}  
    \subitem \textit{Turn off the light.}  

    \item \textbf{Suggest}  
    \subitem Proposes an action, idea, or alternative. May include advice or recommendations.  
    \subitem \textit{Why don’t you try restarting your computer?}  

    \item \textbf{Offer}  
    \subitem Voluntarily provides help, solutions, or resources.  
    \subitem \textit{Would you like some water?}  

    \item \textbf{Reject}  
    \subitem Declines or refuses a proposal, offer, or request. May provide justification or explanation, though this is not required.  
    \subitem \textit{I’m sorry, but I’ll have to pass.}  

    \item \textbf{Encourage}  
    \subitem Provides motivation, praise, or positive reinforcement.  
    \subitem \textit{Don’t worry, you’ll figure it out!}  

    \item \textbf{Manage Topic}  
    \subitem Handles transitions between conversation topics. Can be used for opening, changing, or closing topics.  
    \subitem \textit{Let’s move on to the next point.}  

    \item \textbf{Social Interaction}  
    \subitem Includes greetings and meaningless small talk designed for polite social interaction.  
    \subitem \textit{Hello.}  
    \subitem \textit{How are you?}  
    \subitem \textit{Fine. And you?}  
\end{enumerate}
\section{Human Annotation}
\label{human-annotation}
Across all of our experiments, we employed the help of the following human annotators:
Two native speakers each of English, Chinese, German, and Italian; and one native speaker of Indonesian. One native speaker of English and the native speaker of Indonesian were contributing authors, while the other annotators were recruited by word-of-mouth and compensated between USD \$10--\$15 depending on location. 

The English annotators consisted of one American and one Canadian, both in their 30s. The Chinese annotators were both graduate students. The German annotators were both working professionals. One Italian annotator is retired, and the other is a high school student.

Annotators were informed of the task scope and consented to participate under conditions aligned with ethical research practices. This study was determined to be low-risk and did not require review by an ethics board.

\section{Automated Evaluation of Decoding Back into English}
\label{back-to-english-auto}
We evaluated DAS-decoded English using GPT-4o and GPT-4o-mini, and a direct paraphrase baseline, where the original dialogues were rephrased using a simple paraphrasing prompt\footnote{See Appendix~\ref{para-prompt}}. The paraphrase baseline provides a useful reference point for distinguishing ordinary surface rewording from the more structured transformations introduced by DAS. For example, given the original utterance, ``I’m a bit worried about you going shopping by yourself this afternoon.'' the paraphrased baseline produces ``I'm a little concerned about you heading out to shop alone this afternoon.'' In contrast, DAS decoding generates ``I'm a bit worried about you going shopping alone. Are you sure you'll be okay?'' While the paraphrase baseline makes minor lexical and syntactic adjustments, DAS introduces a more structured transformation by breaking the utterance into multiple turns, adding conversational nuance, or adjusting for different dialogue dynamics.

To ensure robustness and consistency, each model was tested across three runs with a temperature setting of 0.2. To mitigate potential biases, we fixed the encoder and varied the LLM used for DAS decoding, allowing us to assess the effect of different decoding strategies in DAS. The reported scores represent the averages across all runs.

For automated evaluation, we computed BERTScore \citep{bertscore} to measure meaning retention, BLEU \citep{papineni-etal-2002-bleu} to quantify lexical overlap, and ChrF++ \citep{popovic-2015-chrf} to evaluate character-level and word-level similarity between the original and DAS-decoded texts. Since DAS does not use the original sentence as input, we expect the BLEU score to be lower than paraphrasing, while the BERTScore remains high. 
ChrF++ captures both word- and character-level overlap, making it more flexible than BLEU in handling reworded outputs. However, since DAS modifies sentence structure more than standard paraphrasing, we still expect ChrF++ scores to be lower than paraphrasing, reflecting content preservation despite structural variation. 
The results are summarized in Table~\ref{tab:Back2English-results}.

\begin{table}[t]
    \centering
    \small
    \begin{tabular}{lcccc}
    \toprule
        \textbf{Model}  & \textbf{BERTScore} & \textbf{BLEU} & \textbf{ChrF++} \\
        \midrule
        Paraphrasing & 0.943 & 0.184 & 0.389 \\
        GPT4o-mini & 0.909 & 0.126 & 0.343  \\
        GPT4o & 0.914 & 0.142 & 0.369\\
    \bottomrule
    \end{tabular}
    \caption{Semantic (BERTScore) and form-focused (BLEU/ChrF++) similarities between the original and the decoded utterances}
    \label{tab:Back2English-results}
\end{table}

The lower BLEU scores compared to the paraphrase baseline suggest that DAS decoding introduces lexical variety, making it distinct from simple word-for-word reformulation. 
The ChrF++ scores also show that DAS reformulations diverge more from the original structure than direct paraphrasing. Despite this increased divergence, BERTScore remains high (over 0.9, even for the smaller system), reinforcing that DAS effectively preserves intent while rewording the dialogue more flexibly than standard paraphrasing.
The fact that DAS decoding does not have direct access to the original sentence yet still scores relatively close to the paraphrase baseline suggests that its structured encoding influences realization in ways that may limit extreme rewording. Future work could explore whether adjusting encoding constraints allows for more diverse yet meaning-preserving reformulations.

\section{Conversational Context}
\label{appendix-bios}
\begin{table}[t]
\centering
\small
\begin{tabular}{lcccc}
\toprule
    Method & \textbf{Fluency} & \textbf{Coher.} & \textbf{Culture} & \textbf{Situation}  \\
    \midrule
    \textbf{Italian}\\
    \midrule
    Localized & 73 & 70 & 76 & 74  \\
    + Context  & 91 & 85 & 86& 89 \\
    \midrule
    \textbf{German} \\
    \midrule
    Localized  & 82 & 76 & 72 & 76  \\
    + Context  & 89 & 85 & 86 & 89  \\
    \midrule
    \textbf{Chinese} \\
    \midrule
    Localized  & 77 & 78 & 79 & 81  \\
    + Context  & 82 & 80 & 90 & 93  \\
    \bottomrule
\end{tabular}
\caption{Win rates against machine translation and human translation for including a context summary or not.}
\label{table:bio-ablation}
\end{table}

\begin{table*}[t]
    \centering
    \begin{tabular}{l|ccccccc}
    \toprule
      Aspect & DAS Win \% & Both \% & Neither \% & MT Win \% & DAS WR & MT WR & $p$\\
      \midrule
      \textbf{Italian} & & & & \\
      Fluency & 69.2 & 28.5 & \phantom{0}0.0 & \phantom{0}2.3 & 97.7 & 30.8 & <0.001 \\
      Coherence & 82.3 & 14.6 & \phantom{0}0.0 & \phantom{0}3.1 & 96.9 & 17.7 & <0.001\\
      Cultural & 80.8 & 16.2 & \phantom{0}0.8 & \phantom{0}2.3 & 96.9 & 18.5 & <0.001\\
      Situational & 78.5 & 18.5 & \phantom{0}0.0 & \phantom{0}3.1 & 96.9 & 21.5 & <0.001\\
        \midrule
      \textbf{German} & & & \\
      Fluency & 60.8 & 33.8 & \phantom{0}0.8 & \phantom{0}4.6 & 94.6 & 38.5 & <0.001 \\
      Coherence & 56.9 & 34.6 & \phantom{0}0.0 & \phantom{0}8.5 & 91.5 & 43.1 & <0.001\\
      Cultural & 61.5 & 26.2 & \phantom{0}5.4 & \phantom{0}6.9 & 87.7 & 33.1 & <0.001\\
      Situational & 56.9 & 34.6 & \phantom{0}1.5 & \phantom{0}6.9 & 91.5 & 41.5 & <0.001\\
      \midrule
      \textbf{Chinese} & & & \\
      Fluency & 81.1 & 12.1 & \phantom{0}1.5 & \phantom{0}5.3 & 93.2 & 17.4 & <0.001 \\
      Coherence & 80.3 & 10.6 & \phantom{0}0.8 & \phantom{0}8.3 & 90.9 & 18.9 & <0.001\\
      Cultural & 85.6 & \phantom{0}8.3 & \phantom{0}0.8 & \phantom{0}5.3 & 93.9 & 13.6 & <0.001\\
      Situational & 84.1 & 12.1 & \phantom{0}0.8 & \phantom{0}3.0 & 96.2 & 15.2 & <0.001\\
    \end{tabular}
    \caption{Pairwise evaluation of DAS generated conversations vs GPT-4o translated conversations. Ties were classified into ``Both'' and ``Neither''. Win Rate (WR) is Win + ``Both''. $p$-value calculated through two-tailed binomial test}
    \label{tab:das-vs-mt-detailed}
\end{table*}

Early experiments localized and decoded dialogues using DAS alone, without additional conversational context. However, manual inspection and consultation with native speakers revealed room for improvement, particularly in situational appropriateness. The generated dialogues often sounded too formal or stiff in contexts where a more natural or casual tone would have been expected.

One key observation was that nuances such as politeness levels were often lost in the encoding process. This was likely because DAS focuses on extracting content rather than form, whereas politeness and tone are often conveyed through structural and lexical choices rather than explicit meaning. To address this, we incorporated broader conversational context by prompting GPT-4o to generate a summary of the conversation, along with speaker names and biographical details.

Since many languages rely on grammatical gender, we asked GPT-4o to infer or assign speaker genders as part of the biographical information. However, in the initial test, every generated dialogue featured one male and one female character, indicating a bias toward binary gender pairings. To mitigate this, we explicitly modified the prompt to encourage greater diversity in gender assignments.

After this change, the resulting speaker distribution was: 88\% male-female, 6\% male-male (MM), 2\% female-female, 4\% non-binary-female.
Interestingly, for one conversation, a non-binary character was changed into a male character during localization into German and Italian, while remaining non-binary in Chinese. No other characters had gender altered during localization.

The results in Table~\ref{table:bio-ablation} reflect GPT-4o-based evaluation of localized dialogues with and without additional conversational context. While the inclusion of speaker biographies and conversational summaries led to higher GPT evaluation across all criteria, it is important to recognize that GPT-based evaluation may not always align with human judgment. 
Native Italian speakers reviewed a sample of 10 conversations and confirmed GPT's evaluations, suggesting that the inclusion of context genuinely improved fluency, cultural relevance, and situational adaptation. However, given the limited sample size, further human evaluation is required to validate the extent of these improvements across different languages and conversational settings.

\section{DAS vs GPT Translation}
We include the detailed results of human pairwise evaluation of DAS generated conversations compared to GPT-4o translated conversations in Table \ref{tab:das-vs-mt-detailed}.

\section{Human translation vs GPT translation}
\begin{table*}[t]
    \centering
    \begin{tabular}{l|ccccccc}
    \toprule
      Aspect & Human Win \% & Both \% & Neither \% & MT Win \% & Human WR & MT WR & $p$\\
      \midrule
      \textbf{Italian} & & & & \\
      Fluency & \phantom{0}6.9 & 10.0 & \phantom{0}0.8 & 82.3 & 16.9 & 92.3 & <0.001 \\
      Coherence & \phantom{0}9.2 & 14.6 & \phantom{0}0.8 & 75.4 & 23.8 & 90.0 & <0.001\\
      Cultural & \phantom{0}7.7 & 18.5 & 23.1 & 50.8 & 26.2 & 69.2 & <0.001\\
      Situational & \phantom{0}3.8 & 36.2 & \phantom{0}3.1 & 56.9 & 40.0 & 93.1 & <0.001\\
        \midrule
      \textbf{German} & & & \\
      Fluency & \phantom{0}4.6 & 13.8 & \phantom{0}1.5 & 80.0 & 18.5 & 93.8 & <0.001 \\
      Coherence & \phantom{0}5.4 & 36.9 & \phantom{0}1.5 & 56.2 & 42.3 & 93.1 & <0.001\\
      Cultural & \phantom{0}3.1 & 30.0 & \phantom{0}1.5 & 65.4 & 33.1 & 95.4 & <0.001\\
      Situational & \phantom{0}4.6 & 21.5 & \phantom{0}1.5 & 72.3 & 26.2 & 93.8 & <0.001\\
      \midrule
      \textbf{Chinese} & & & \\
      Fluency & 10.0 & 41.1 & \phantom{0}4.4 & 44.4 & 51.1 & 85.6 & <0.001 \\
      Coherence & 14.4 & 46.7 & \phantom{0}5.6 & 33.3 & 61.1 & 80.0 & 0.014\\
      Cultural & \phantom{0}7.8 & 50.0 & 11.1 & 31.1 & 57.8 & 81.1 & <0.001\\
      Situational & 11.1 & 34.4 & \phantom{0}5.6 & 48.9 & 45.6 & 83.3 & <0.001\\
    \end{tabular}
    \caption{Pairwise evaluation of human translated conversations vs GPT-4o translated conversations (MT). Ties were classified into ``Both'' and ``Neither''. Win Rate (WR) is Win + ``Both''. $p$-value calculated through two-tailed binomial test}
    \label{tab:human-vs-mt-detailed}
\end{table*}

We analyzed the difference in human perceptions of GPT4o translated conversations and professionally translated conversations. The annotators were not shown the original conversations, so translation accuracy is not taken into consideration for this experiment. We found that for cultural relevance there was a high number of ``Neither'' answers for Italian and Chinese, but not German. The machine translation versions of the conversations were significantly scored higher than the professional human translations, which may have prioritized fidelity to the source material over the generation of a fluent and natural dialogue. The Chinese dialogues were the most competitively scored, either indicating a higher quality of the human translations or a challenge for the LLM.  Detailed results are shown in Table \ref{tab:human-vs-mt-detailed}.
\section{Prompts}
\label{prompts}
\subsection{Paraphrase}
\label{para-prompt}
Produce a new conversation from the given dialogue by paraphrasing each utterance.\\
\\
Conversation:\\
<conversation>

\subsection{Machine Translation}
\label{mt-prompt}
Translate the following conversation into <language>.\\
\\
Conversation:\\
<conversation>\\

\subsection{Single Prompt Localize+Translate}
\label{ltprompt}
Translate the following conversation into <language>. While translating, please localize the dialogue for <language> speakers. This should include any necessary changes to names, locations, social dynamics, common objects (replace any brands or items with more commonly used ones), and general cultural appropriateness to make the context feel natural for <language> speakers. Assign culturally appropriate names based on gender, age, and relationship dynamics in the target culture. Be mindful of specifying politeness levels, family dynamics, and relevant cultural norms.
\\
Conversation:\\
<conversation>\\

\subsection{Encode}
You will read dialogue snippets. Assign a function label to each utterance with all necessary parameters to reconstruct the meaning. The goal is to capture what the speaker is doing (e.g., asking a question, making a request, giving feedback) rather than how they say it.  The 'parameters' of the functions will be whatever is necessary to capture the meaning of the utterance.  This should be the minimum amount of information necessary to convey all of the information of the sentence.\\
\\
Here is the complete list of functions with descriptions and examples:\\
\\
<function name>: <description>\\
    - example: <example>\\
...\\
\\
Note: It's possible for one utterance (or even one sentence) to serve multiple purposes.  In this case, it's fine to choose more than one, but keep them in the order presented.\\
Example: \\
text: ``No, I don't think so'',\\
functions: [``disagree()'', ``express(doubt)'']\\
\\
Conversation:\\
<conversation>\\

\subsection{Generate Context}
Summarize the scene by creating details about the characters to capture the context of the dialogue.  If a name is provided, use that, but if not, feel free to make up details.  Don't use the same names as the example.  Provide at minimum, each speaker's name, gender (M,F,X), age, and presumed relationship to the other speaker.  Try to capture the context of the scene.  Don't let every conversation be between a man and a woman.  Try to vary up the gender combinations.\\
\\
Example:\\
Two coworkers, Alex (M, 35) and Jamie (X, 28), are discussing a project deadline and planning next steps. Alex is a project manager, Jamie is a software developer.  The conversation takes place in the office break room, where they often chat about after-work activities.\\
\\
Conversation:\\
<conversation>\\

\subsection{Localize Context}
You will be provided with a scenario in which a dialogue is taking place.  Please localize the dialogue context for <language> speakers. This should include any necessary changes to names, locations, social dynamics, common objects (replace any brands or items with more commonly used ones), and general cultural relevance to make the context feel natural for <language> speakers. Assign culturally appropriate names based on gender, age, and relationship dynamics in the target culture. Be mindful of specifying politeness levels, family dynamics, and relevant cultural norms.\\
Do NOT write a sample conversation.  Only provide the localized scenario.\\
\\
Scenario:\\
<context>\\
\\
Target language/culture: <language>\\

\subsection{Localize DAS}
Please localize the following Dialogue Act Script for <language> speakers while maintaining the original structure and meaning. Do not remove, condense, or add new topics. Only adjust cultural references when necessary, and keep all turns intact. The format must remain exactly the same, with only localized modifications where relevant.\\
\\
Target language/culture: <language>\\
Summary: <localized context>\\
\\
DAS:\\
<DAS turns>\\
\\
\subsection{Decode}
You are given a conversation setting with details about the speakers, their ages, genders, and relationships. Use this information to generate the text of the conversation based on the provided functions for each turn. Consider the speakers' ages, relationships, and any relevant details to make the conversation natural and contextually accurate. It is okay to leave out or make up parts of the functions if they don't fit what the characters would naturally say. Aim for cultural authenticity even if the names of the characters/places/foods need to be changed.\\
\\
You don't have to stick to one function per sentence.  Some functions will combine naturally into a single sentence.\\
Example: \\
functions: A.disagree(); A.express(doubt)\\
A: ``No, I don't think so''\\
\\
Do not merge multiple turns into a single response. Maintain the same turn structure. Ensure that each turn corresponds to an individual line of dialogue. Do not repeat or shorten any of the functions or dialogue history.\\
\\
Language: <language>\\
Context: <localized context>\\
Conversation:\\
<localized DAS turns>
\end{document}